%% file: main.tex
\documentclass{article}
\input{math_commands.tex}

\usepackage{microtype}
\usepackage{graphicx}
\usepackage{subfigure}
\usepackage{booktabs} 
\usepackage{cancel}

\usepackage{hyperref}
\usepackage{url}
\usepackage{nicefrac}
\usepackage{lipsum}
\usepackage{xcolor}

\usepackage{arxiv}

\usepackage{amsmath}
\usepackage{amssymb}
\usepackage{mathtools}
\usepackage{amsthm}
\usepackage[svgpath=Figures/]{svg}
\usepackage[capitalize,noabbrev]{cleveref}

\theoremstyle{plain}

\theoremstyle{definition}

\theoremstyle{remark}

\usepackage[textsize=tiny]{todonotes}

\title{One-Shot Federated Learning with Bayesian Pseudocoresets}

\author{
Tim d'Hondt\\
Eindhoven University of Technology\\
\texttt{t.dhondt@tue.nl} \\
\And
Mykola Pechenizkiy\\
Eindhoven University of Technology\\
\texttt{m.pechenizkiy@tue.nl}\\
\And
Robert Peharz\\
Graz University of Technology\\
\texttt{robert.peharz@tugraz.at}
} 
\begin{document}
\maketitle

\begin{abstract}
Optimization-based techniques for federated learning (FL) often come with prohibitive communication cost, as high dimensional model parameters need to be communicated repeatedly between server and clients. 
In this paper, we follow a Bayesian approach allowing to perform FL with \textit{one-shot communication}, by solving the global inference problem as a product of local client posteriors.
For models with multi-modal likelihoods, such as neural networks, a naive application of this scheme is hampered, since clients will capture different posterior modes, causing a destructive collapse of the posterior on the server side.
Consequently, we explore approximate inference in the function-space representation of client posteriors, hence suffering less or not at all from multi-modality. 
We show that distributed function-space inference is tightly related to learning Bayesian pseudocoresets and develop a tractable Bayesian FL algorithm on this insight. 
We show that this approach achieves prediction performance competitive to state-of-the-art while showing a striking reduction in communication cost of up to two orders of magnitude.
Moreover, due to its Bayesian nature, our method also delivers well-calibrated uncertainty estimates.
\end{abstract}
\input{Sections/Introduction}
\input{Sections/Background}
\input{Sections/Method2}
\input{Sections/RelatedWork}
\input{Sections/Experiments}
\input{Sections/Conclusion}

\bibliographystyle{plain}
\bibliography{references}

\newpage
\appendix
\onecolumn
\input{Sections/Appendix}
\end{document}

%% file: math_commands.tex
\usepackage{amsmath,amsfonts,bm}

\def\eqref#1{equation~\ref{#1}}

\def\1{\bm{1}}

\def\rvtheta{{\mathbf{\Theta}}}

\def\rvf{{\mathbf{f}}}

\def\rvx{{\mathbf{x}}}
\def\rvy{{\mathbf{y}}}

\def\vtheta{{\bm{\theta}}}

\def\vf{{\bm{f}}}

\def\vu{{\bm{u}}}

\def\vx{{\bm{x}}}
\def\vy{{\bm{y}}}
\def\vz{{\bm{z}}}

\def\mX{{\bm{X}}}

\def\mZ{{\bm{Z}}}

\DeclareMathAlphabet{\mathsfit}{\encodingdefault}{\sfdefault}{m}{sl}
\SetMathAlphabet{\mathsfit}{bold}{\encodingdefault}{\sfdefault}{bx}{n}



%% file: Sections/Introduction.tex
\section{Introduction}
\label{sec:intro}
Federated learning (FL) is a machine learning setting where multiple data-owning entities (\textit{clients}) collaborate on solving a global learning problem (usually coordinated by a \textit{server}) without sharing their raw data \cite{kairouz2021FLsurvey}. 
For large neural networks, gradient-based federated optimization (FO) methods are the de-facto standard class of algorithms \cite{wang2021fieldguide}. 
These methods minimize a loss function over partitioned data using a variation of distributed gradient descent, where in each step (\textit{communication round}) the server shares the current global model parameters and clients return the gradient of their local likelihood function. For models requiring thousands of server gradient steps until convergence, the \emph{communication cost} of FO can become prohibitive, as it scales proportionally to the number of communication rounds, model size and number of clients. 
A popular heuristic to reduce the amount of communication is federated averaging (FedAvg) \cite{mcmahan2017communication}, i.e.~to perform several local gradient updates per communication round before returning a \emph{pseudo-gradient}, i.e.~the difference between their updated parameters and the current server parameters \cite{mcmahan2017communication, reddi2021adaptivefo}. 
While empirically effective, this heuristic is known to cause \textit{client drift}, a phenomenon where the server model converges to a non-optimal solution due to the biased pseudo-gradients \cite{scaffold2020, charles2021convergencetradeoffs}. 

Alternatives to optimization-based techniques are Bayesian approaches to FL, formulating learning as a distributed inference problem, where the global server posterior over parameters is proportional to the model prior and the product of local client likelihoods \cite{al-shedivat2021federated, ashman2022partitioned}. 
Besides delivering model uncertainties, Bayesian inference also allows---in principle---a \emph{one-shot approach} to FL, i.e.~using only a single communication round.
Specifically, each client approximates its local model posterior, sends it to the server, which subsequently aggregates these 'local factors' into a global approximate posterior. 
If the client posteriors are exact, this approach will also deliver the exact global posterior, i.e.~the same posterior as if all client data had been combined into a single dataset.
Hence, Bayesian FL does in principle not suffer from systematic biases such as client drift.  

However, when dealing with modern neural networks, a central problem of Bayesian FL is that likelihood functions are highly non-linear and hard to approximate. 
Moreover, it is well known that the parameter space of neural nets has many symmetries, such as the ones generated by exchanging two neurons in the same layer. 
These symmetries result in non-identifiability of the model and a huge number of equivalent local minima (in an optimization point of view) or posterior modes (in a Bayesian view). 
Meanwhile, existing Bayesian FL methods typically restrict local factors to simple, unimodal functions, often from the exponential family (e.g \cite{ashman2022partitioned, guo2023federated, al-shedivat2021federated}).
Attempting one-shot Bayesian FL with these techniques will almost certainly fail, since each local factor will capture a mode in a different region of the parameter space, causing the global posterior density to collapse to zero where support is not shared over all clients \cite{ashman2022partitioned, de2022parallel}. 
Therefore, these methods still require multi-round algorithms combined with heavy damping to periodically align the supports of local factors in the parameter space, again increasing the communication burden. 
Also switching to more expressive posterior representation, such as generative models, seems futile due to the sheer number of equivalent modes in the parameter posterior.

\textbf{Function Space FL.} 
The core obstacle to one-shot FL illustrated above is non-identifiability of model parameters, or in other words, that the relationship between model parameters and represented functions is 'many-to-one'.
The problem disappears when we adopt a \emph{function-space} view, that is, when we treat the Bayesian neural network (BNN) as a stochastic process \cite{burt2021understanding, rudner2020rethinking}.
In function space, the posterior will concentrate on a single function in the infinite data limit (when restricting to the support of the data), i.e., the model is identifiable in function-space, except for pathological cases.
Note that formally the function-space posterior is given by the push-forward of the weight-space posterior, reducing all equivalent ``symmetric'' modes into one mode in function-space.
Thus, working in function space is an amenable way to perform communication-efficient one-shot Bayesian FL. 

\textbf{Bayesian Pseudocoresets for FL.}
While the function-space view illustrated above is elegant, representing and communicating the local function-space posteriors is non-trivial, especially for neural networks. 
We propose that each client learns and shares a small set of key function values that serve as an approximate sufficient statistic of their data, similar to inducing points in sparse Gaussian processes \cite{snelson2005sparse,burt2021understanding}. When combined with the functional prior, these function values induce a posterior that resembles the true local posterior in function space and automatically captures all equivalent symmetrical modes in weight space.

We show that learning inducing points in the function space of neural networks is tightly related to learning a \emph{Bayesian pseudocoreset} (BPC) \cite{manousakas2020, Kim2022BPCDiv}. 
In a nutshell, a BPC is a collection of synthetic data points whose \emph{weight} posterior approximates the weight of the entire dataset. 
Their interpretation as (approximate) inducing points in function space inference is perhaps not surprising, but, to the best of our knowledge, has not been deeply studied yet.
This connection motivates to leverage recent advances in BPC learning for a tractable Bayesian FL algorithm in function space, using one-shot communication between clients and server.
Specifically, each client learns a BPC on their private data, which is then sent to the server.
The server then trains a model on the combined coresets of all clients, either in a Bayesian manner (e.g.~using Monte Carlo) or via standard gradient-based optimization.

\begin{figure}[t]
\centering
\begin{subfigure}
    \centering
    \includegraphics[width=0.23\textwidth]{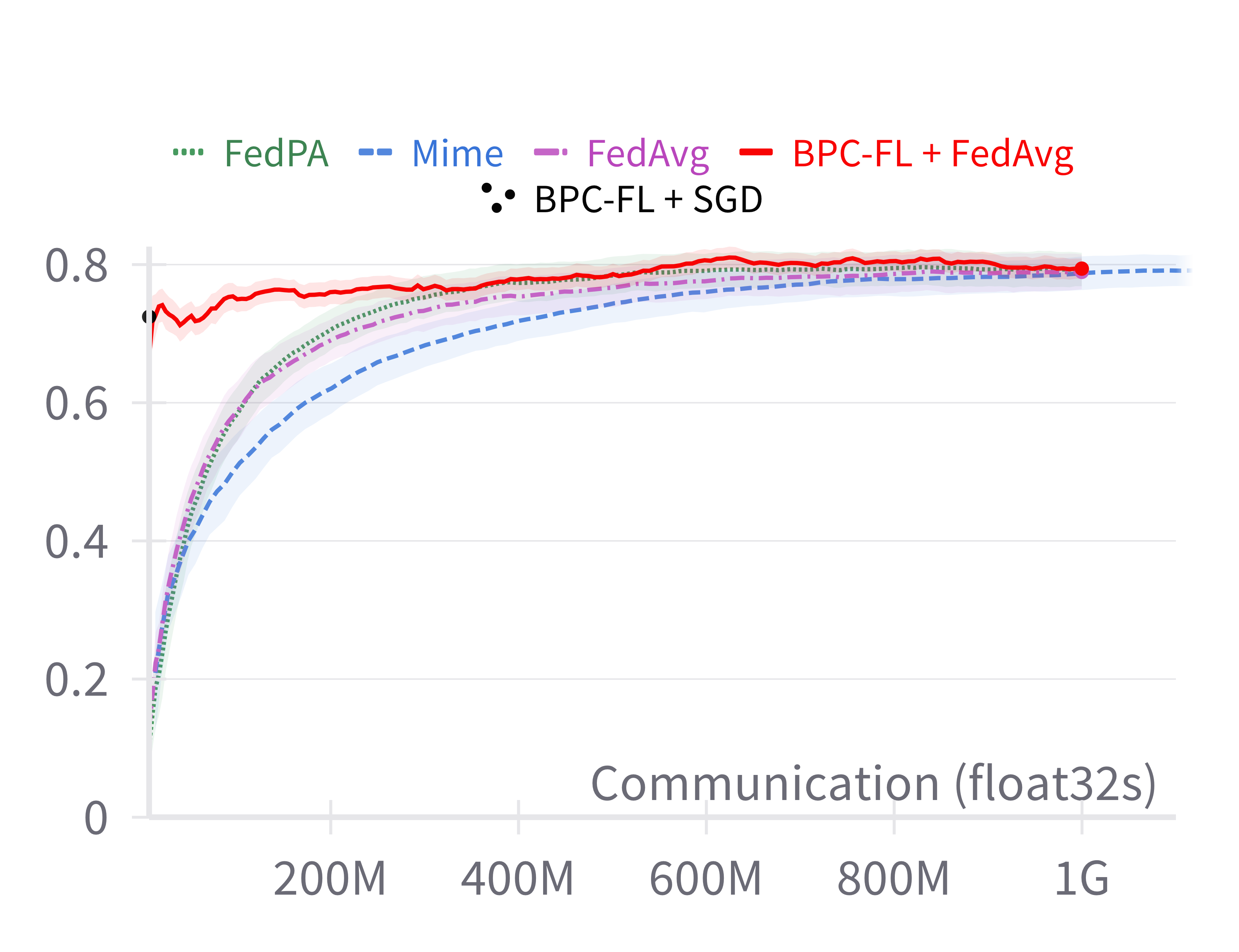}
\end{subfigure}%
\begin{subfigure}
    \centering
    \includegraphics[width=0.23\textwidth]{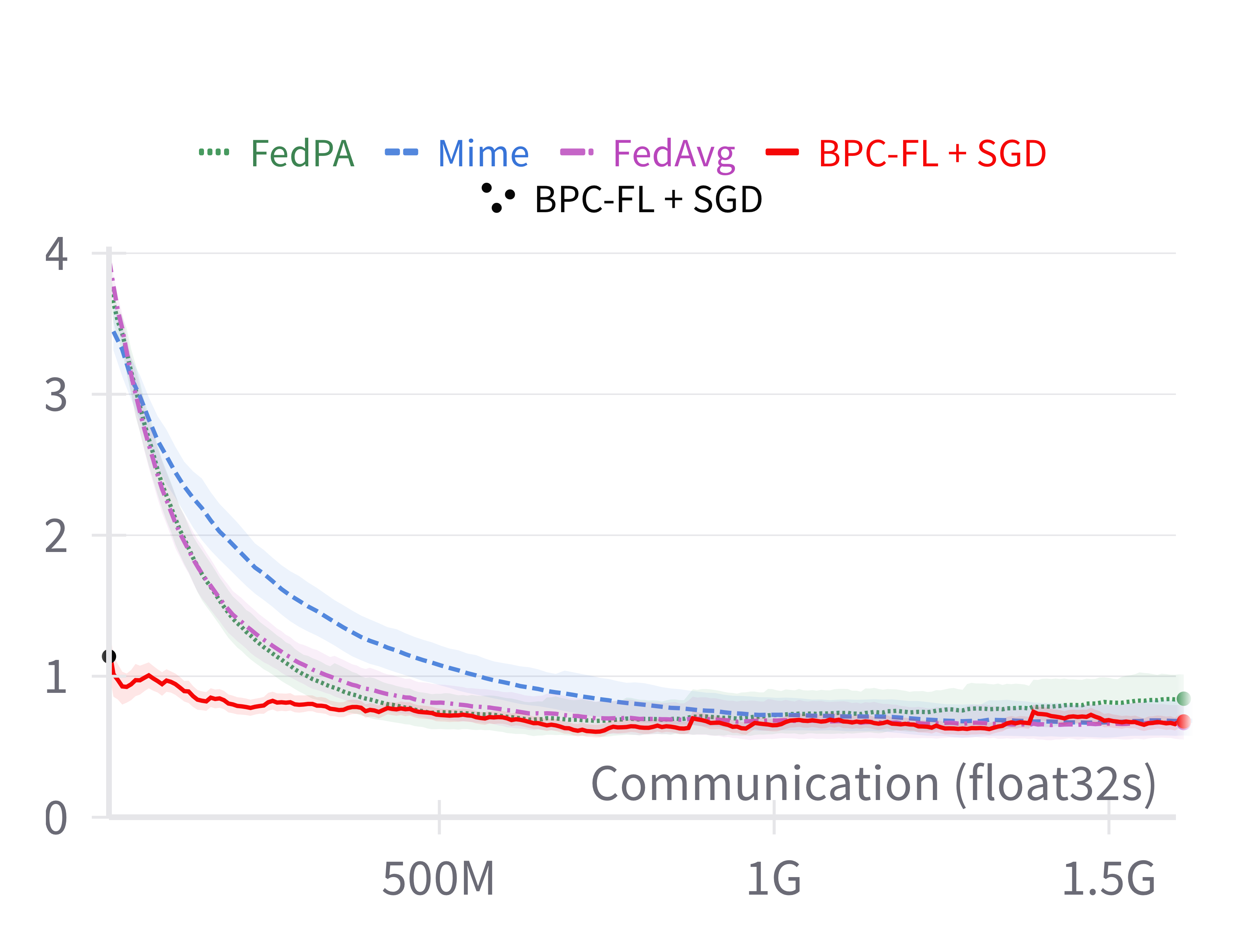}
\end{subfigure}
\begin{subfigure}
    \centering
    \includegraphics[width=0.23\textwidth]{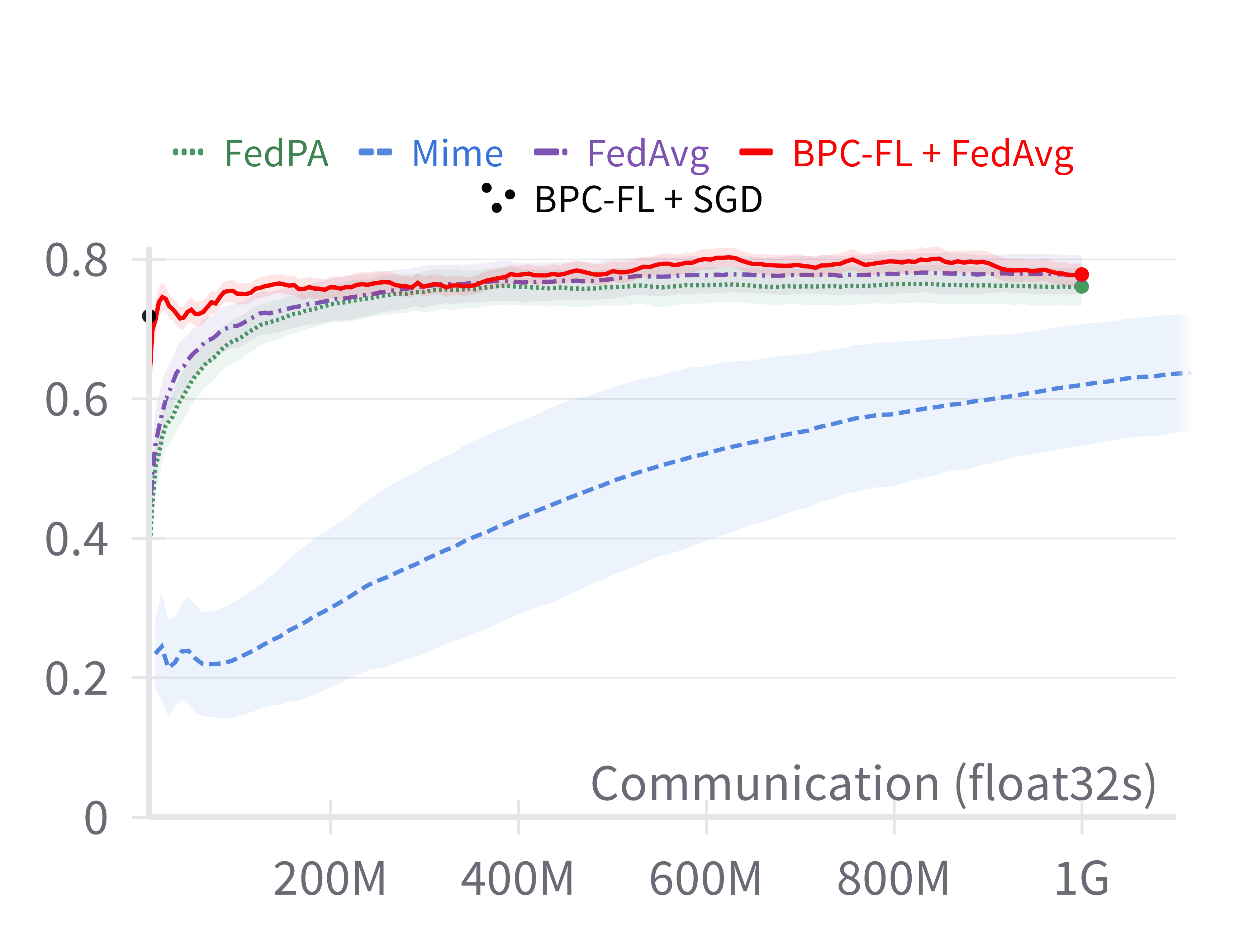}
\end{subfigure}%
\begin{subfigure}
    \centering
    \includegraphics[width=0.23\textwidth]{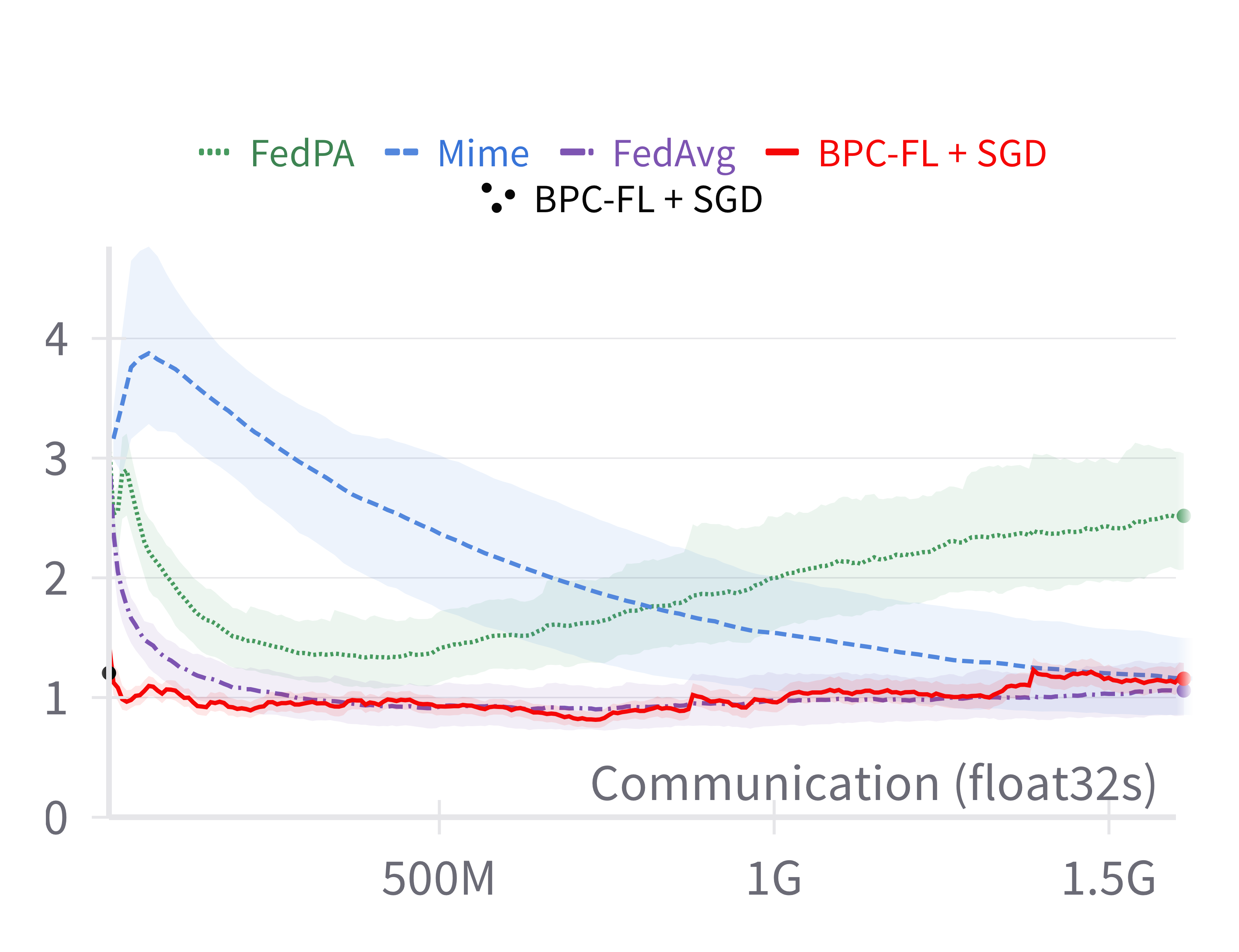}
\end{subfigure}
\caption{Results on EMNIST-62. Left column shows accuracy (y-axis) vs. communication in float32s (x-axis). Right column shows negative log likelihood (y-axis) vs communication in float32s (x-axis) Upper row shows baselines with 10 local steps, bottom row with 100 local steps. Note our one-shot approach (\textbf{black} $\bullet$) in the upper left corner and our one-shot approach as FedAvg initialization in \textbf{\textcolor{red}{red}}.}
\label{fig:emnistconvergence}
\end{figure}

The result of this simple idea is striking, as illustrated in Figure~\ref{fig:emnistconvergence}, showing performance as a function of communication for our method and various competitors, all run on the same neural architecture and the EMNIST-62 benchmark \cite{caldas2018leaf}.
After just a single communication round, needing less than $7M$ single floats communicated, our method achieves $72\%$ classification accuracy.
This is competitive to the best comparison method, which achieves ultimately slightly less than $80\%$, but \emph{requires hundreds of communication rounds.}
In total, the most communication-efficient competitor required at least $119M$ single floats communicated to match our performance of $72\%$.
\emph{Thus, our method is between one and two orders of magnitude more communication efficient than state-of-the art.}

Moreover, when using our solution as initialization for \mbox{FedAvg}, the \emph{resulting method dominates all other methods for any amount of communication spent.}
This advantage is in particular evident in the low communication regime (i.e.~one- or few-shot FL).
Besides being orders of magnitude more communication-efficient than weight-space FL approaches, our approach also has the benefits of being completely asynchronous, producing epistemic uncertainty estimates and not suffering from client drift. 

In summary, our main contributions are:
\begin{itemize}
\item We propose a novel approach for ultra communication-efficient one-shot learning of BNNs in the FL setting by locally fitting and communicating inducing points in the function-space representation.
\item We show how this function-space inference approach is well approximated by distributed learning of BPCs.
\item We propose BPC-FL, a first tractable algorithm for performing function-space inference in the FL setting.
\item In empirical evaluations we demonstrate that our method is competitive to state-or-the-art FL approaches, while using \emph{up to two orders of magnitude less communication,} and produces models with favorable uncertainty estimates.
\end{itemize}

%% file: Sections/Background.tex
\section{Preliminaries}
We consider conditional probabilistic models for a random vector $\rvy \in \mathcal{Y}$, conditioned on $\rvx \in \mathcal{X} \subseteq \mathbb{R}^D$, where $\mathcal{Y} \subseteq \{0, 1\}^C$ for classification and $\mathcal{Y} \subseteq \mathbb{R}^C$ for regression tasks. Let $\rvtheta \subseteq \mathbb{R}^P$ be a random vector containing the model parameters of a Bayesian Neural Network (BNN) with an architecture defined by $f: \mathcal{X} \times \mathbb{R}^P \rightarrow \mathbb{R}^C$. 
Hence, $f$ is a random function that for a specific parameter realization $\vtheta$ of $\rvtheta$ encodes a mapping from $\mathcal{X}$ to a set of $C$ logits. Given a dataset $\mathcal{D} = \{y_n, x_n\}^N_{n=1} = (\mX, \vy)$ of $N$ observations of $\rvy$ and $\rvx$, we assume a likelihood $p(\vy|f(\mX, \vtheta)) = \prod^N_{n=1} p(y_n|f(x_n, \vtheta))$. Letting $p_0$ be a prior density over $\rvtheta$, the posterior density for $\rvtheta = \vtheta$ is given as:
\begin{equation}\label{eq:vanillaposterior}
    p(\vtheta|\mathcal{D}) \propto p_0(\vtheta)p(\vy|f(\mX, \vtheta))
\end{equation}
which has an intractable normalization constant $Z(\mathcal{D}) = \int_{\mathbb{R}^P} p_0(\vtheta)p(\vy|f(\mX, \vtheta))d\vtheta$ due to $f$ being a nonlinear function in $\vtheta$.

In the federated learning (FL) setting, we assume the  dataset $\mathcal{D}$ is partitioned over $M$ data-generating entities \textit{(clients)}: $\mathcal{D} = \mathcal{D}_1 \cup ... \cup \mathcal{D}_M$ that collaborate in learning a model under the supervision of a \textit{server}, while any function of a client's dataset $\mathcal{D}_m$ of size $n_m$ can only be evaluated locally. These local datasets $\mathcal{D}_m$ may be \textit{non-iid} in the sense that each $\mathcal{D}_m$ is sampled from a (slightly) different distribution.

\subsection{Federated Posterior Inference}
\label{sec:foandfpi}

As described in the Section \ref{sec:intro}, the default FL algorithm is \textit{federated optimization} (FO), which is a version of distributed gradient descent and communication hungry.
Federated averaging (FedAvg) \cite{mcmahan2017communication} aims to save on communication, but the introduction of pseudo-gradients also introduces a systematic bias denoted as client drift \cite{scaffold2020, charles2021convergencetradeoffs}.


Alternatively, one might formulate FL as Bayesian inference problem.
The \textit{federated posterior inference problem (FPI)} is the task of computing the following posterior within the restrictions of the FL setting: 
\begin{equation}\label{eq:federatedposterior}
    p(\vtheta|\mathcal{D}) \propto p(\vtheta) \prod^M_{m=1} p(\vy_m|f(\mX_m, \vtheta)),
\end{equation}
where $p(\vy_m|f(\mX_m, \vtheta))$ is the $m^{\text{th}}$ client likelihood.
If the client likelihoods (or equivalently the local posteriors divided by the prior density) could be communicated exactly, it is evident that (\ref{eq:federatedposterior}) is identical to (\ref{eq:vanillaposterior}), hence, in principle FPI does not introduce a systematic bias such as client drift.
Even when using approximate likelihoods, a certain robustness of FPI regarding client drift has been empirically observed \cite{al-shedivat2021federated}. 

Since the likelihoods can only be evaluated locally, the usual approach is to compute $p(\vtheta|\mathcal{D})$ with a \textit{message passing algorithm}, which infers a target density using a collection of localized inferences with an approximate density $q(\vtheta) \propto p(\vtheta) \prod^M_{m=1} t_m(\vtheta),$
where $t_m(\vtheta)$ are called the \textit{local factors}, or \textit{site parameters} \cite{pearl1986fusion, vehtari2020expectation}. The local factors are then typically fitted to minimize the reverse KL divergence \cite{bui2018partitioned} or its forward substitute \cite{hasenclever2017distributed, vehtari2020expectation, guo2023federated}. Both objectives have been applied to the FPI problem \cite{guo2023federated, ashman2022partitioned}, with the factors $t_m(\vtheta)$ restricted to be in the exponential family. 

Since these client factors $t_m(\vtheta)$ are implemented with a unimodal approximation, they might been drawn to different modes of the likelihood in local inference, especially in the case of non-identifiable models such as neural networks. 
With high probability, this will cause catastropic failures when aggregating the factors at the server as $q(\vtheta) \propto p(\vtheta) \prod^M_{m=1} t_m(\vtheta)$.
Thus, $q(\vtheta) \approx 0$ at any $\vtheta$ where even just one factor has close to zero probability \cite{de2022parallel}. 
Therefore, these methods still require many communication rounds and heavy damping of the server updates to converge, especially when performing local updates at clients in parallel \cite{ashman2022partitioned, guo2023federated}. 

\subsection{Function space representation of BNNs}

The posterior collapse problem described above occurs in non-identifiable models, such as BNNs.
A natural remedy is to switch to a 'function space' view, by regarding the function $f(., \rvtheta)$ as a stochastic process on the domain $\mathcal{X}$.
Intuitively, $f(., \rvtheta)$ is an infinite dimensional random vector indexed by $\mathcal{X}$ that for a particular parameter instantiation $\vtheta$ encodes to what logit values a NN architecture maps all possible elements of $\mathcal{X}$. Let $\rvf$ denote the random vector $f(., \rvtheta)$ and $\rvf_X$ denote that vector on a subset of the domain $X \subseteq \mathcal{X}$. For brevity, we will also denote $\rvf$ its value for a specific parameter $\vtheta$ as $\rvf^\vtheta$, such that $f(\mX, \vtheta) = \rvf^\vtheta_\mX$. Given a prior probability density $p_0$ over $\rvtheta$, we can define a function-space prior and posterior distribution as follows \cite{wolpert1993bayesian, rudner2022tractable}:
\begin{align}
    p_0(\rvf) &= \int_{\mathbb{R}^P} p_0(\vtheta)\delta(\rvf - \rvf^{\vtheta}) d\vtheta\label{eq:fnspaceprior} \\
    p(\rvf|\mathcal{D}) &= \int_{\mathbb{R}^P} p(\vtheta|\mathcal{D})\delta(\rvf - \rvf^{\vtheta}) d\vtheta \label{eq:fnspaceposterior}
\end{align}
where $\delta$ is the Dirac delta function. These densities can also be defined more generally as pushforward measures through the tools of measure theory \cite{burt2021understanding}, but we will stick to the definition above, without loss of generality. It is noticeworthy that the KL divergence between two distributions over functions generated from different distributions over parameters applied to the same neural network architecture is well-defined if the KL divergence between the distributions over parameters is finite, by the strong data processing inequality \cite{polyanskiy2017strong}:
\begin{equation}\label{eq:bpsobjective}
    \mathbb{D}_{KL}(p(\rvf^\vtheta)||q(\rvf^{\vtheta})) \leq \mathbb{D}_{KL}(p(\vtheta)||q(\vtheta))
\end{equation}
As a consequence, it is possible to perform variational inference in the function space of neural networks instead of their parameter space  \cite{sun2018functional, burt2021understanding, rudner2022tractable}. 

\subsection{Bayesian Pseudocoresets}\label{sec:bayesianpseudocoresets}
Computing or even estimating the BNN posterior (\ref{eq:vanillaposterior}) is in general computationally expensive, in particular for large $N$.
\emph{Bayesian pseudocoresets} (BPCs) introduce the idea to replace the real data with a small synthetic dataset $\mathcal{C} = \{\vz_k, \vu_k\}^K_{k=1} = (\mZ, \vu)$ of \textit{pseudolabels} $\vu_k \in \mathcal{Y}$ and \textit{pseudoinputs} $\vz_k \in \mathcal{X}$ where $K \ll N$, which produces the \textit{coreset posterior} \cite{manousakas2021}:
\begin{align}\label{eq:coresetposterior}
    q(\vtheta|\mathcal{C}) &\propto p_0(\vtheta)p(\vu|f(\mZ, \vtheta))\\
    &= p_0(\vtheta) \exp \left(\sum^K_{k=1} - \log p(\vu_k|f(\vz_k, \vtheta)) \right)
\end{align}
Typically, $\mathcal{C}$ is trained to minimize some divergence measure $\mathbb{D}$ between the coreset posterior $q(\vtheta|\mathcal{C})$ and the target posterior $p(\vtheta|\mathcal{D})$:
\begin{equation}\label{eq:bpcobjective}
    \mathcal{C}^* = \textrm{arg } \underset{\mathcal{C}}{\min}\, \mathbb{D}\left[p(\vtheta|\mathcal{D}), q(\vtheta|\mathcal{C})\right]
\end{equation}
Intuitively, a BPC is a summary of the information about $\vtheta$ that is contained in $\mathcal{D}$, i.e.~an \textit{approximate sufficient statistic}. 
Thus, $\mathcal{C}$ compresses $\mathcal{D}$ by exploiting redundancy in the data with respect to $f$ \cite{winter2023machine}. 

BPC approximations of posteriors come with some favorable properties over other approximate posterior families, such as the fact that they preserve important model structure that is exhibited in the true posterior (symmetries, subspace structures, heavy tails) through $f$, which makes it an interesting choice for modeling posteriors involving complex likelihood functions. Moreover, BPC posteriors are trivially composable through set operations on the coresets and are inference algorithm agnostic at downstream tasks \cite{winter2023machine}.
Different choices for the divergence measure $\mathbb{D}$ and approaches to minimize that measure give rise to different BPC algorithms \cite{Kim2022BPCDiv, kim2023function, manousakas2020, manousakas2021}.
BPC learning is also closely related to dataset distillation and dataset condensation \cite{Cazenavette_2022_CVPR, zhao2021dataset} as shown in \cite{Kim2022BPCDiv}.

Intuitively, since BPCs have a particular reminiscence to inducing points in Gaussian processes (GPs) \cite{snelson2005sparse,burt2021understanding}, one might wonder whether they can be interpreted as a sparse function-space inference method.
In the following section we show that this indeed the case and exploit them for a one-shot Bayesian FL algorithm.

%% file: Sections/Method2.tex
\section{Bayesian One-Shot Federated Learning}

\subsection{Federated Learning in Function-space}
For a partitioned dataset over clients $\mathcal{D} = \mathcal{D}_1 \cup ... \cup \mathcal{D}_M$, let us denote $\mathcal{D}_m = (\mX_m, \vy_m)$ for each client $m$. Also, let us split up $\rvf_\mX$ into $\rvf_\mX = [\rvf_{\mX_1}, ..., \rvf_{\mX_M}]$, denoting the \emph{latent function values} indexed by the input values in each client's dataset. Further, we denote $\rvf_*$ to be the latent function values at all other inputs in the domain $\mathcal{X}$. We can then write the federated posterior from equation (\ref{eq:federatedposterior}) in its function space representation as follows:
\begin{align}
    p(\rvf_*, \rvf_\mX|\mathcal{D}) &= \frac{p_0(\rvf_*, \rvf_\mX)}{p(\mathcal{D})}\prod^M_{m=1}p(\vy_m|\rvf_{\mX_m})\\
    &= \int_{\mathbb{R}^P} p(\vtheta|\mathcal{D})\delta(\rvf_\mX^\theta - \rvf_\mX)\delta(\rvf_*^\theta - \rvf_*) d\vtheta \nonumber 
\end{align}

Instead of estimating the weight space posterior $p(\vtheta|\mathcal{D})$ within the restrictions of the FL setting, we will take an approach inspired by inducing points in sparse GPs \cite{rasmussen2003gaussian, snelson2005sparse, titsias2009variational},
which is---under particular modeling assumptions depicted below---equivalent to learning BPCs.

Let us assume a set of function values $\rvf_\mZ = [\rvf_{\mZ_1}, ..., \rvf_{\mZ_M}]$ where $\mZ_m$ are \emph{pseudo inputs} belonging to client $m$. In our model, each variable $\rvf_{\mZ_m}$ is treated as a \emph{sufficient statistic} of the client's local data function values $\rvf_{\mX_m}$, that is, we make the prior assumption that $\rvf_\mX$ and $\rvf_*$ are conditionally independent given $\rvf_{\mZ_m}$ and all $\rvf_{\mX_m}$ are conditionally independent given their local $\rvf_{\mZ_m}$:
\begin{equation}
p(\rvf) = p_0(\rvf_\mZ)p(\rvf_*|\rvf_\mZ)\prod^M_{m=1}p(\rvf_{\mX_m}|\rvf_{\mZ_m})    
\end{equation}
where $\rvf = [\rvf_\mX, \rvf_\mZ, \rvf_*]$ for brevity of notation.
If the dimensionality $K$ of each $\rvf_{\mZ_m}$ is significantly smaller than the client's dataset size, i.e.~$K \ll n_m$, then $\rvf_{\mZ_m}$ is a compressed representation of the information about $\rvf_{\mX_m}$ in local dataset $\mathcal{D}_m$. 
In contrast to sparse GPs, where pseudo inputs $\mZ_m$ are often fixed, we let clients optimize these, additionally to $\rvf_\mZ$.
Including the likelihood terms, the full joint model over latent $\rvf$ and observed $\vy$ is given as:
\begin{equation}
    p(\rvf, \vy) = p(\rvf)\prod^M_{m=1} p(\vy_m|\rvf_{\mX_m})
\end{equation}

Note that in the FL setting, it is difficult for a client $m$ to perform inference on $\rvf_{\mZ_m}$ given $\mathcal{D}_m$, because $\rvf_{\mZ_m}$ is dependent on all other $\rvf_{\mZ_m}$ and thus other client data through prior $p_0(\rvf_Z)$. 
One solution would be to iteratively refine the posterior density over each $\rvf_{\mZ_m}$ in an EP-type inference scheme.
In order to pursue the goal of one-shot FL, however, we make the assumption that all $\rvf_{\mZ_m}$ are independent: $p_0(\rvf_\mZ) = \prod^M_{m=1} p_0(\rvf_{\mZ_m})$, resulting in the following posterior (see Appendix \ref{sec:appendix:derivation_indep_posteriors} for a derivation):
\begin{align}\label{eq:fnspacefedposterior}
    p(\rvf|\mathcal{D}) = p(\rvf_*|\rvf_{\mZ})\prod^M_{m=1}p(\rvf_{\mZ_m}, \rvf_{\mX_m}| \vy_m)
\end{align}
Through our assumptions, we have now divided the problem of computing the global posterior $p(\rvf|\mathcal{D})$ into a set of $M$ independent local posterior computations of $p(\rvf_{\mZ_m}, \rvf_{\mX_m}|\rvy_m)$, which can then be communicated to a server and used to compute the predictive distribution $p(\rvf_*|\rvf_{\mZ})$ in downstream tasks. 

\subsection{Approximate Inference}
The next question is: how do we locally compute and communicate the intractable posteriors $p(\rvf_{\mZ_m}, \rvf_{\mX_m}|\rvy_m)$? We choose to fit an approximate posterior $q(\rvf_{\mZ_m}, \rvf_{\mX_m}) = q(\rvf_{\mZ_m})p(\rvf_{\mX_m}|\rvf_{\mZ_m})$ at each client in parallel by minimizing the forward KL divergence $\mathbb{D}_{KL}\left[p(\rvf_{\mZ_m}, \rvf_{\mX_m}|\vy_m)||q(\rvf_{\mZ_m}, \rvf_{\mX_m})\right]$ with respect to $q$. To be able to apply gradient-based solvers, $q$ is required to be a differentiable, parametric density function.

Another important insight is that we want to incorporate the functional prior $p_0(\rvf_{\mZ_m})$ into our definition of $q$, as this allows us to model the complex dependencies between all function values in $\rvf_{\mZ_m}$ through the model architecture. Following this insight, we choose $q(\rvf_{\mZ_m})$ to be a posterior given some learnable \textit{pseudo-labels}, such that $q(\rvf_{\mZ_m}) = p(\rvf_{\mZ_m}|\hat{\vy}_m) \propto p_0(\rvf_{\mZ_m})p(\hat{\vy}_m|\rvf_{\mZ_m})$. The client local objective then looks as follows by following the proof from Proposition 3.1 in \cite{kim2023function}, which is based on \cite{matthews2016sparse} and ignoring terms that are independent of $\mZ_m$ and $\hat{\vy}_m$:
\begin{align}\label{eq:localobjective}
    \mathcal{L}_m(\mZ_m, \hat{\vy}_m) &= \mathbb{D}_{KL}\left[p(\rvf_{\mZ_m}, \rvf_{\mX_m}|\vy_m)||q(\rvf_{\mZ_m}, \rvf_{\mX_m})\right] \nonumber \\ 
    &= \mathbb{E}_{p(\rvf_{\mZ_m}|\vy_m)}\left[\log \frac{p(\hat{\vy}_m|\rvf_{\mZ_m})}{p(\hat{\vy}_m|\mZ_m)} \right]
\end{align}
Where $p(\hat{\vy}_m|\mZ_m)$ is the marginal likelihood of $(\hat{\vy}, \mZ_m)$. Notice from equation \ref{eq:federatedposterior} that we can sample from $p(\rvf_{\mZ_m}|\vy_m)$ using the Law of the Unconscious Statistician, i.e monte carlo sampling $\tilde{\vtheta} \sim p(\vtheta|\mathcal{D})$ and mapping it into the function $\rvf^{\tilde{\vtheta}}_{\mZ_m} = f(\mZ, \tilde{\vtheta})$ through the network architecture. The objective then simplifies to $\mathcal{L}_m(\mZ_m, \hat{\vy}_m)= \mathbb{E}_{p(\vtheta|\mathcal{D}_m)}\left[\log \frac{p(\hat{\vy}_m|\rvf^\vtheta_{\mZ_m})}{p(\hat{\vy}_m|\mZ_m)} \right]$, which is the BPC objective when choosing $\mathbb{D}$ to be the forward KL divergence in equation \ref{eq:bpcobjective}. 

Thus, our parameterization of $q(\rvf_{\mZ_m})$ via pseudo-labels $\hat{\vy}_m$, can be seen as approximate function space inference for learning a BPC $(\mZ_m, \hat{\vy}_m)$.
This allows us to leverage recent advances in BPC learning to tractably minimize (\ref{eq:localobjective}) for large neural networks. 
This 'pseudo-label' parameterization of an approximate posterior over function space inducing points $q(\rvf_{\mZ_m}) \propto p_0(\rvf_{\mZ_m})p(\hat{\vy}_m|\rvf_{\mZ_m})$ has also been used in the Sparse Variational Gaussian Process (SVGP) literature \cite{adam2021dual}.

Our motivation for using the forward KL divergence is that this divergence forces $q(\rvf_{\mZ_m})$ to have a broad support over all functions that are probable under $p(\rvf_{\mZ_m}|\mathcal{D}_m)$, which is important for server aggregation and offsets some of the bias that we introduced by assuming an independent prior $p_0(\rvf_\mZ)$. Moreover, contrary to the reverse KL divergence, choosing the forward KL divergence prevents us from having to backpropagate through a Monte Carlo sampling process that estimates the expectation in equation \ref{eq:localobjective}, which makes the procedure more scalable \cite{Kim2022BPCDiv}.

\subsection{BPC-FL}
In this section, we propose a first FL algorithm that solves the defined inference problem, which we will call \textbf{BPC-FL}. The general idea is simple and summarized in Algorithm \ref{alg:bpcfl}: In parallel, each client $m$ learns his BPC $(\mZ_m, \hat{\vy}_m)$ by minimizing equation (\ref{eq:localobjective}) and finally communicating $q(\rvf_{\mZ_m})$ represented as $(\mZ_m, \hat{\rvy}_m)$ to the server. The server then simply concatenates the BPCs as $\mZ = [\mZ_1, ..., \mZ_M]$ and $\hat{\rvy} = [\hat{\vy}_1, ..., \hat{\vy}_M]$ to obtain $q(\rvf_\mZ) = p(\rvf_\mZ|\hat{\vy})$, which can then be used to compute $p(\rvf_*|\vf_\mZ)$ in downstream tasks.

\textbf{Learning the local pseudodata}
Having shown the connection between BPC learning and equation \ref{eq:localobjective}, we apply the BPC-fKL algorithm from \cite{Kim2022BPCDiv} to minimize equation \ref{eq:localobjective} locally at each client $m$ with respect to his pseudodata, which we denote in short as $\mathcal{C}_m = (\mZ_m, \hat{\vy}_m)$. This choice of BPC inference algorithm is orthogonal to our general one-shot FL approach, so performance of our proposed method is expected to increase if new, better BPC inference algorithms are proposed over time.

The gradient of the local objective $\mathcal{L}_m$ with respect to $\mathcal{C}_m = (\mZ_m, \hat{\vy}_m)$ is given as follows \cite{Kim2022BPCDiv}:
\begin{equation}\label{eq:klgradient}
    \nabla_{\mathcal{C}}\mathcal{L}_m = \mathbb{E}_{p(\vtheta|\mathcal{D}_m)}\left[\nabla_\mathcal{C}\log p(\hat{\vy}_m|f(\mZ_m, \vtheta)) \right] + \mathbb{E}_{p(\vtheta|\mathcal{C}_m)}\left[- \nabla_\mathcal{C}\log p(\hat{\vy}_m|f(\mZ_m, \vtheta)) \right]
\end{equation}
and can then be plugged into any gradient-based solver of choice. The expectations over $p(\vtheta|\mathcal{D}_m)$ and $p(\vtheta|\mathcal{C}_m)$ are estimated with short-chain MCMC sampling, to make the gradient estimator similar to the one proposed in the Contrastive Divergence algorithm \cite{hinton2002training}. The BPC-fKL algorithm applies a number of tricks to the estimation of (\ref{eq:klgradient}) to improve performance and reduce computation and memory cost. Most importantly, it pretrains MAP optimization trajectories on $p(\vtheta|\mathcal{D})$ and saves checkpoints of weights to disk, which can be loaded during training as initialization points for the MCMC sampler \cite{Cazenavette_2022_CVPR}. We share the initialization seeds for these trajectories among clients to ensure similar spread of sampling from $p(\vtheta|\mathcal{D})$ throughout the parameter space. For completeness, we provide the pseudocode for BPC-fKL in Appendix \ref{sec:appendix:bpc-fklpseudocode}.

\textbf{Initialization of $\mZ$ and $\hat{\vy}_m$}
We initialize each $\vz \in \mZ_m$ randomly from a Gaussian centered at the element-wise mean $\bar{\mX}_m$ of a client $m$'s data: $\vz \sim \mathcal{N}(\bar{\mX}_m, \sigma_{\vz}) \hspace{2mm} \forall \vz \in \mZ_m$, with $\sigma_{\vz}$ being a hyperparameter to determine the initialization noise. For classification tasks, we initialize pseudolabels $\hat{\vy}_m$ in order by cycling through the unique labels in a client's dataset, sorted by their occurence. For regression tasks, we initialize $\hat{\vy}_m$ from the dataset average output $\vy_m$. 

\textbf{Server aggregation}
Constructing the server posterior $q(\rvf_\mZ) = \prod^M_{m=1}q(\rvf_{\mZ_m})$ is as simple as concatenating all received $\mZ_m$ and $\vy_m$ vectors. In experiments, we found that it is helpful to weigh each $q(\rvf_{\mZ_m})$ proportional to $\frac{n_m}{N}$, to put more emphasis on $q(\rvf_{\mZ_m})$ that carry information from more data.

\begin{algorithm}[H]
    \centering
    \caption{BPC-FL}\label{alg:bpcfl}
    \begin{algorithmic}[1]
        \STATE {\bfseries Input:} local data $\mathcal{D}_1, ..., \mathcal{D}_M$
        \FOR{$m \in \{1, ..., M\}$ in parallel}
        \STATE $(\mZ_m, \hat{\vy}_m) \leftarrow$ BPC-fKL$(\mathcal{D}_m)$ \cite{Kim2022BPCDiv}
        \STATE Send $(\mZ_m, \hat{\vy}_m)$ to server
        \ENDFOR
        \STATE $\mZ \leftarrow [\mZ_1, ..., \mZ_M], \hat{\vy} \leftarrow [\hat{\vy}_1, ..., \hat{\vy}_M]$
        \STATE {\bfseries{Return}  $q(\rvf_\mZ)$}
    \end{algorithmic}
\end{algorithm}

\subsection{Downstream inference tasks}
After completing Algorithm \ref{alg:bpcfl}, the server has access to $q(\vf_\mZ)$, which provides end users with more flexibility in what downstream tasks we can tackle than standard FL approaches.

One of the most important downstream tasks will be to predict a set of unseen test datapoints $\mX_*$. This can be done by estimating the predictive distribution $\mathbb{E}_{q(\vf_\mZ)}(\rvf_*|\rvf_\mZ) = \int_{\mathbb{R}^P} p(\vtheta|\mathcal{C}) \delta(\rvf_*^\vtheta - \rvf_*)d\vtheta$. We have total freedom in whether we want to obtain a MAP estimate from $p(\vtheta|\mathcal{C})$ with for example SGD or whether we want to obtain samples from it using Hamiltonian Monte Carlo.

Thinking a little more creative, BPC-FL could also be used to enhance existing FL methods. For example, we could find a mode of $q(\vtheta|\mathcal{C})$ and use it as an initializer for an FO algorithm as a 'jump start' to save costly communication rounds, something which we show in our experimental results in Figure \ref{fig:emnistconvergence}. 



%% file: Sections/RelatedWork.tex
\section{Related work}
Formulating the goal of federated learning as a Bayesian inference problem as in equation (\ref{eq:federatedposterior}) was first proposed in \cite{al-shedivat2021federated}, where the authors estimated Gaussian local factor approximations to un-bias client pseudo-gradients used in an FO algorithm, but did not learn a posterior on the server side. Solving a distributed posterior inference problem with a product of local factors has a long history in Expectation Propagation (EP) \cite{vehtari2020expectation, hasenclever2017distributed}, Bayesian Committee Machines (BCM) \cite{tresp2000bayesian} and embarrassingly parallel VI \cite{neiswanger2015embarrassingly, neiswanger2014asymptotically}, but was first mentioned in the context of FL for Partitioned Variational Inference (PVI) \cite{ashman2022partitioned}. \cite{guo2023federated} then adapted PVI to instead minimize the forward KL divergence. All of these works use a unimodal approximation for the local factors, which fails for one-shot inference as mentioned in \cite{ashman2022partitioned, de2022parallel}. An alternative approach is to estimate the global posterior by collecting samples using distributed MCMC algorithms fitted to the FL setting \cite{el2021distributed, vono2022qlsd, karagulyan2023elf, chen2021fedbe}, though these methods require many communication rounds, similar to FO algorithms.

Learning a model in the FL setting in one communication round (one shot) has been proposed through model distillation at the server in \cite{chen2021fedbe, guha2019one, li2020practical, hasan2024calibrated}, though these works assume a publicly available dataset at the server. As a response, Zhang et al. \cite{zhang2022DENSE} replaced the public dataset with a generative model.

Another approach to one-shot FL is assume a publicly available feature extractor and learn a classification head by having each client send a function-space distribution for each class \cite{beitollahi2024parametric}.

A number of works exist that achieve one-shot FL with dataset distillation; also through sharing synthetic datasets with the server \cite{zhou2020distilled, goetz2020federated, hu2022fedsynth, xiong2023feddm}. Dataset distillation techniques are actually closely related to Bayesian Pseudocoresets \cite{Kim2022BPCDiv}, but typically use optimization instead of MCMC sampling.

Formulating FL as a function space inference problem has been proposed in \cite{hutchinsonfunctionspace}, though in the context of performing federated variational inference on the parameter prior in a hierarchical Bayesian model, and with only preliminary empirical results. In \cite{dhawan2023leveraging}, it is proposed to perform the server aggregation step of an FO algorithm in function space using estimations of local Fisher information matrices, in order to reduce client drift. 

%% file: Sections/Experiments.tex
\section{Experimental Results}
In this section we compare the predictive performance and uncertainty estimates of models produced by our method and models trained by other federated learning methods. Because communication efficiency is the main motivation for this paper, we pay special attention to the amount of communication required to obtain a certain result for each method.

\subsection{Experimental Setup}
\textbf{Datasets and model architecture}
Empirical results are obtained from learning tasks on two synthetic datasets and a large FL benchmark dataset from the LEAF repository \cite{caldas2018leaf}. The synthetic datasets; one regression problem and one 2d classification problem, were chosen mainly to evaluate the uncertainty estimates of models and to get an intuitive view on what the pseudodata at the server looks like. To evaluate predictive performance on relatively large and complex FL learning tasks, we use the EMNIST-62 dataset from LEAF that is split across clients as to carry natural non-iid-ness \cite{caldas2018leaf, reddi2021adaptivefo}. MLP architectures are applied to the synthetic dataset problems and a ConvNet architecture is selected for EMNIST-62 based on the one used in \cite{reddi2021adaptivefo}. Details can be found in Appendix \ref{sec:appendix:datasets}. 

\textbf{Baselines}
We compare against FedAvg \cite{mcmahan2017communication} and two FO algorithms that are designed to prevent client drift: MIME \cite{karimireddy2021breaking} and FedPA \cite{al-shedivat2021federated}. We test these methods using different amounts of local updates, in order to improve their communication efficiency, but also to evaluate the expected reduction in performance due to client drift. For the synthetic datasets, we also compare against Distilled One Shot Federated Learning (DOSFL); a method that also promises one-shot federated learning, but using dataset distillation \cite{zhao2021dataset}. Because DOSFL requires backpropagation through an entire gradient descent trajectory, we were not able to compute a baseline for this algorithm on EMNIST-62 due to memory limitations. Hyperparameters of baseline methods were set according to their best performing configuration reported in their respected paper, if available. These and other details can be found in Appendix \ref{sec:appendix:baselines}.

\textbf{Evaluation metrics}
To evaluate the predictive performance of our method, we apply different algorithms on $p(\vtheta|\mathcal{C})$ to produce neural network weights. We use SGD and Adam to compute MAP estimates of $p(\vtheta|\mathcal{C})$ and use HMC \cite{neal2011mcmc} to produce samples that we use in a Monte Carlo estimation of the Bayesian predictive $\mathbb{E}_{p(\vtheta|\mathcal{C})}[p(y'|f(x', \vtheta))]$ of a test example $(x', y')$. 
We also run SGD on $p(\vtheta|\mathcal{C})$ and use the resulting MAP estimate as an initializer for FedAvg, with the goal of having FedAvg converge in less communication rounds by giving it a 'head start' through sending $\mathcal{C}$ first.
Details and hyperparameters for these downstream samplers and optimizers are given in Appendix \ref{sec:appendix:eval}.

For EMNIST-62, we record negative log likelihoods and accuracies on a test set that is produced by withholding 20 percent of each client's data from training. We also record intermediary estimates of these 2 metrics after each communication round on a random sample of clients, as to visualize convergence of FO methods. To evaluate the uncertainty calibration of models, we report the expected calibration error (ECE) \cite{naeini2015obtaining} for each method. Finally, we keep track of the communication cost of each method as the amount of single floats that are communicated from and to a (fictive) server, to evaluate the communication cost - performance trade off. All metrics are reported are an average over 5 different seeds.

\textbf{Implementation details}
All experiments on the benchmark tasks were conducted using an adapted version of FedJAX \cite{ro2021fedjax}. Author's implementations were used for FedAvg and MIME, whereas FedPA and DOSFL were implemented by us.

\subsection{Main results}
\begin{figure}[t]
\begin{center}
\centerline{\includegraphics[scale=0.15]{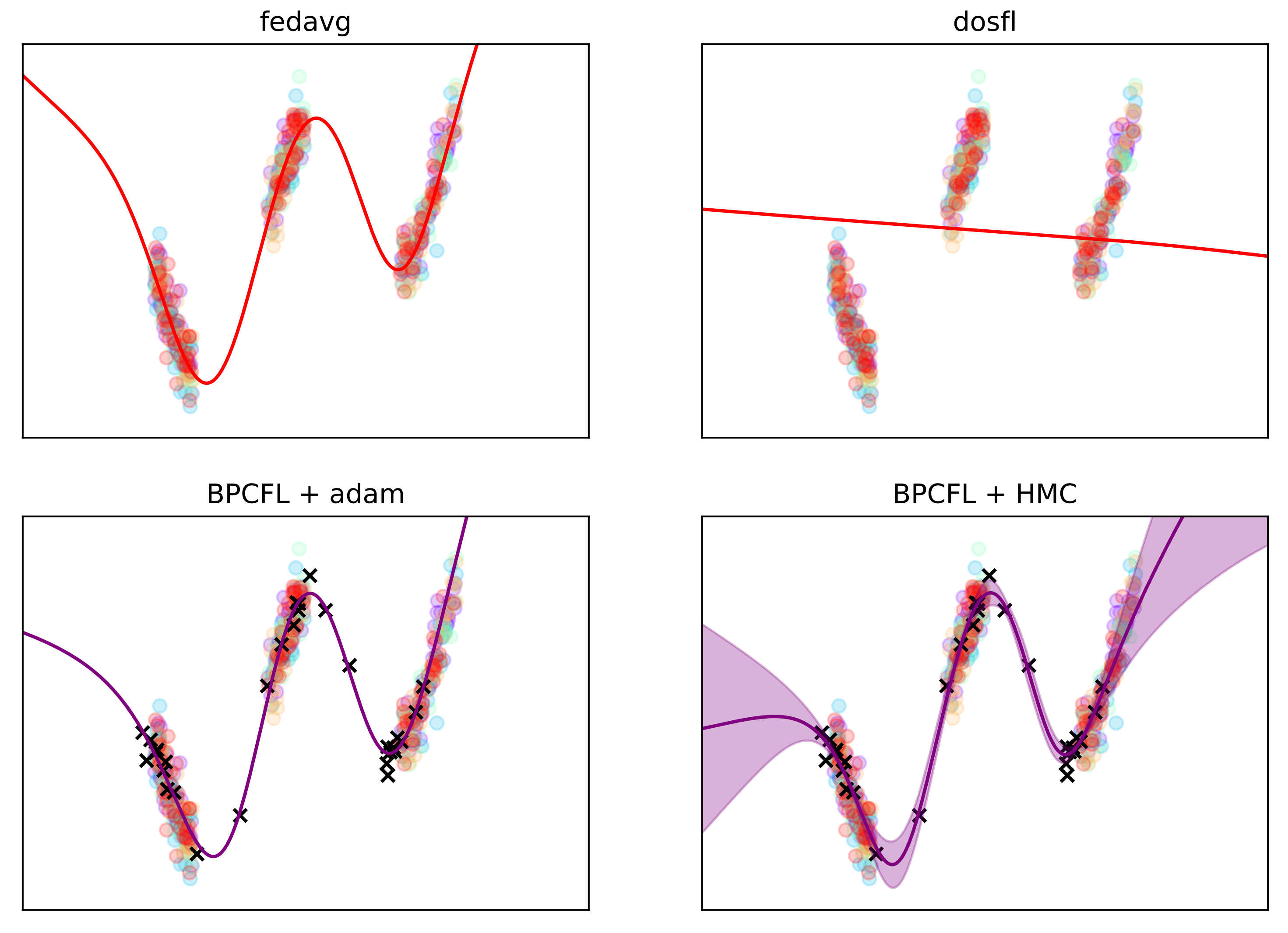}}
\caption{Synthetic regression: client data is visualized in a different color for each client. Upper row presents the produced server models by FedAvg (left) and DOSFL (right). Lower row shows $\mZ$ as black crosses. The purple line left shows a model learned by running Adam on $q(\vtheta|\mathcal{C})$, whereas the purple line and area show the mean and one standard deviation values for the Bayesian predictive distribution estimated by HMC samples from $q(\vtheta|\mathcal{C})$.}
\label{fig:regressionresults}
\end{center}
\end{figure}

\textbf{Synthetic regression}
Figure \ref{fig:regressionresults} shows the predictive distributions of a model using weights that are learned by FedAvg, DOSFL and our method BPC-FL on a synthetically generated regression dataset that was partitioned over 5 clients. Each client has 100 datapoints and learns a pseudodataset of size $K=6$ in BPC-FL, which are then sent to the server. Concluding from Figure \ref{fig:regressionresults}, BPC-FL seems to produce predictive distributions on the inducing points that generalize well to the entire dataset. This is especially noticeworthy from a communication cost perspective when considering that FedAvg requires around 100 communication rounds to converge to a good result with 10 local update steps, whereas BPC-FL only requires 6 one dimensional pseudodatapoints to be sent to the server in one shot. DOSFL was not tested on a regression task in the original paper and therefore does not produce good results on this task for any hyperparameter setting. More results on smaller client datasets can be found in Appendix \ref{sec:appendix:additional}, where it is shown that FedAvg starts to overfit, but BPC-FL keeps producing good results.

\begin{figure}[t]
\begin{center}
\centerline{\includegraphics[scale=0.30]{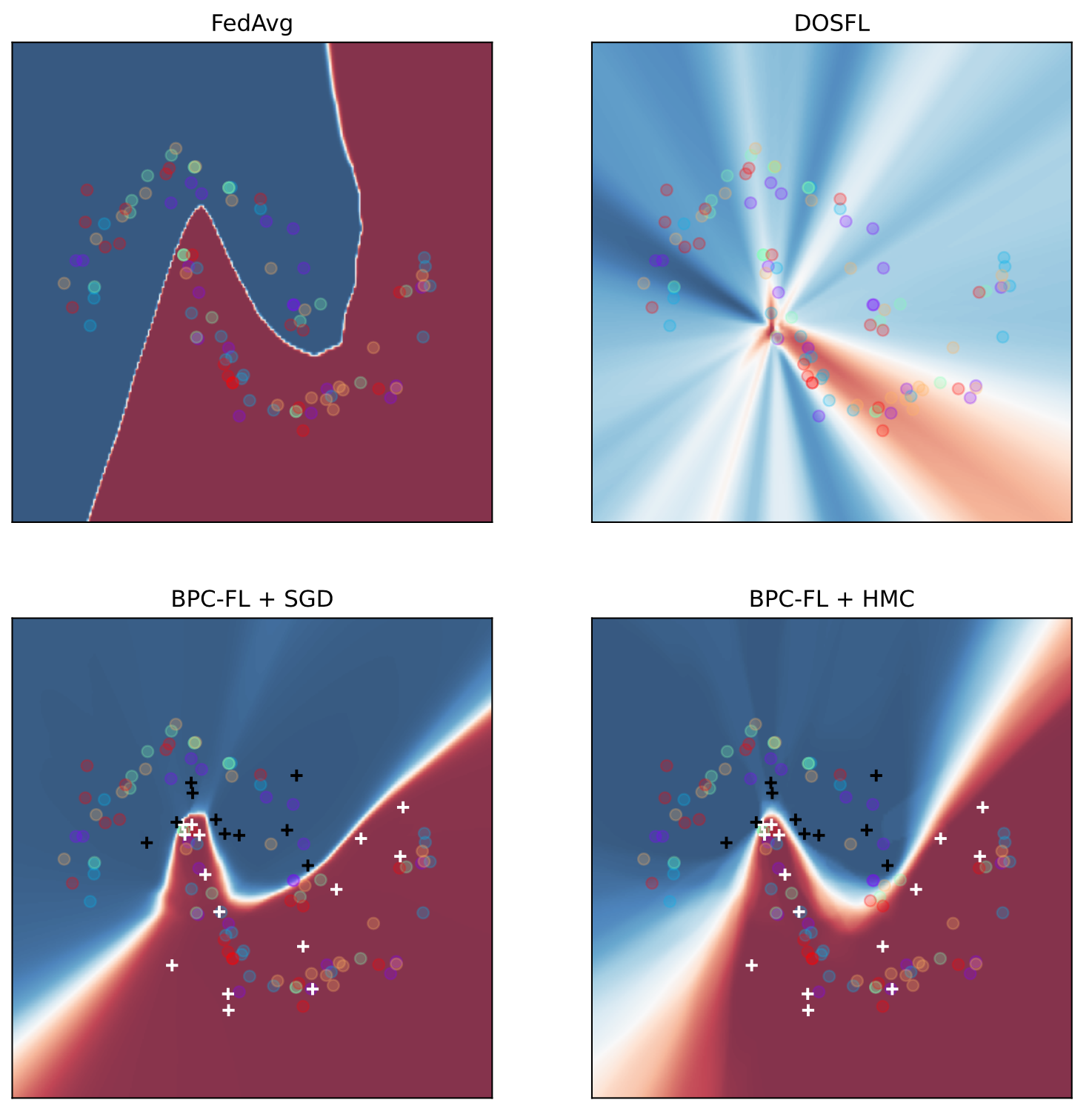}}
\caption{2d classification: red-blue filling shows each method's predictive distribution. On the bottom row, the crosses represent $\mZ$ and their color represents $\hat{\vy}_m$.}
\label{fig:2dclassificationresults}
\end{center}
\end{figure}

\textbf{2d classification}
Figure \ref{fig:2dclassificationresults} shows the predictive distributions of models produced by the methods. With a pseudodataset of size only 5, BPC-FL with SGD produces a decent predictive distribution and BPC-FL with HMC produces a predictive distribution that estimates the epistemic uncertainty on the full dataset very well. For points that were intitialized randomly from a Gaussian distribution, $\mZ$ converges to a logical solution. After an extensive hyperparameter search, the performance for DOSFL is again disappointing.

\begin{table*}[t]
\centering
\begin{tabular}{l|l|l|l|l|l}
                      & \multicolumn{2}{l|}{\textbf{Negative log likelihood}} & \multicolumn{2}{l|}{\textbf{Accuracy}} & \textbf{ECE} \\ \hline
\textbf{Method}       & Final       & Communication       & Final         & Communication   &                    \\ \hline
FedAvg  (10 local steps)              &   $1.038 \pm 0.035$          & $264e^6$                          &   $0.791 \pm 0.006$            &        $250e^6$                      &          $0.246 \pm 0.035$          \\
FedPA   (10 local steps)              &   $1.538 \pm 0.083$          &        $244e^6$                   &    $0.790 \pm 0.006$           & $216e^6$                             &  $0.252 \pm 0.031$                  \\
MIME    (10 local steps)              &    $ 2.254 \pm 0.113 $       &   $435e^6$                        &    $0.775 \pm 0.010$           &   $398e^6$                            &      $0.261 \pm 0.033 $             \\
FedAvg (100 local steps) &    $1.703\pm0.058$         &   $138e^6$           &     $0.775\pm 0.007$          &         $119e^6$                     &  $0.303\pm0.04$                  \\
FedPA (100 local steps) &     $4.960 \pm 0.259$        &    -                       &   $ 0.768 \pm 0.005 $          &               $150e^6$              &      $0.264 \pm 0.076$             \\
MIME (100 local steps)  &    $1.471\pm0.461$        &    $1545e^6$                       &      $0.722\pm0.014 $        &              $3043e^6$                &       $0.167 \pm 0.083$             \\ \hline
BPC-FL + SGD (ours)  &    $1.203 \pm 0.004$         &     $6.6e^6$                      &       $0.723 \pm 0.003$        &     $6.6e^6$                         &    $0.086 \pm 0.020$                \\
BPC-FL + Adam (ours) &      $1.750 \pm 0.001$       &      -                     &   $0.696 \pm 0.010$            &      -                        &      $0.028 \pm 0.003$              \\
BPC-FL + HMC (ours)  &     $1.349 \pm 0.011$        &       -                    &    $0.675 \pm 0.001$           &              -                &      $0.139 \pm 0.010$             \\ \hline
\end{tabular}
\label{tbl:emnist62results}
\caption{Negative log likelihood (NLL), accuracy and ECE results for EMNIST-62. For NLL and accuracy, we report both the final metrics and the average amount of communication (in number of floats) it takes to match the performance of BPC-FL + SGD. Standard deviation of a metric over different seeds is also given.}
\end{table*}

\textbf{EMNIST-62}
For this larger FL benchmark, we mostly focus on evaluating the trade-off between performance and communication cost. We consider here a scenario with a pool of 100 training clients and a pseudodataset size of 62. Metrics are reported over test clients that were not in the training pool. Considering the final negative log likelihood and accuracy in Table \ref{tbl:emnist62results}, we see that BPC-FL cannot achieve the final performance of FedAvg with 10 local steps, but does actually beat all baselines on negative log likelihood when we increase the amount of local steps, likely because the baseline methods start to suffer from client drift. Also, we observed in experiments that the test performance of FO methods started to decrease over time, an indication of overfitting. While BPC-FL does not achieve state-of-the-art final performance, we can always run SGD on $p(\vtheta|\mathcal{C})$, use this as an initialization for any FO algorithm and still obtain FO algorithm final performance, though at a fraction of the communication cost, as is shown in Figure \ref{fig:emnistconvergence} by the red line.

In terms of communication cost though, BPC-FL obtains its competitive predictive results with orders of magnitude less communication, concluding from Table \ref{tbl:emnist62results}. FedAvg with 10 local steps requires on average 40 times more communication to achieve the same performance as BPC-FL with SGD as a downstream algorithm. This clearly demonstrates that this novel approach of communicating function-space inducing points has a lot of potential in reducing the communication cost of FL. Finally, we also see that BPC-FL produces models that have better calibrated uncertainty estimates, as shown by the Expected Calibration Error.
We provide additional results showing the influence of the amount of participating clients and the amount of inducing points on predictive performance in Appendix \ref{sec:appendix:additional}.

%% file: Sections/Conclusion.tex
\section{Conclusions}
In this paper, we explored doing FL in a single communication round by performing local approximate inference on a set of inducing points in the function space representation of neural networks. We showed empirically that this approach has potential to decrease communication cost of FL by orders of magnitude, while still being able to produce server models with good predictive performance and favorable uncertainty estimates over existing FL algorithms.

\textbf{Limitations}
A clear limitation at this point is the computation and memory cost of BPC-fKL for learning the client BPCs. It requires a very large amount of disk space to store the pretrained optimization trajectories and has in our experience a 10-100 fold longer runtime over e.g FedAvg. Also, with BPC-fKL being based on Contrastive Divergence \cite{hinton2002training}, it is notoriously instable and difficult to tune. We expect our method can be significantly improved once the scientific field of BPC inference (which is still in its infancy) matures and produces better and more efficient algorithms.

Another limitation is the lack of privacy guarantees or clear compatibility with contemporary privacy-guaranteeing techniques. Differential privacy guarantees have been shown to be easily implemented in BPC learning \cite{manousakas2020} through using DP-SGD and leveraging the post-processing property of differential privacy. However, since the fKL-BPC algorithm saves many intermediate training states to disk and uses these states throughout the optimization of the BPC, the sensitivity of a BPC to training data is non-trivial, as is the sensitivity of the final server BPC to an individual or an entire client dataset. In general, adding privacy guarantees to dataset distillation (a specific instance of BPC \cite{Kim2022BPCDiv}) is an ongoing and controversial topic \cite{dong2022privacy, carlini2022no}, and this paper merely serves as a motivating use-case for more work to be done in this direction. Nonetheless, we consider the proposed method as an interesting change of perspective on how to approach distributed inference or federated learning of neural networks with extreme communication efficiency.

\section{Acknowledgements}
This project was partially supported by KPN SmartTwo and H2020 SmartChange. We also thank SURF (www.surf.nl) for the support in using the National Supercomputer Snellius. 

%% file: Sections/Appendix.tex
\section{Derivations}
\subsection{Derivation of independent posterior}\label{sec:appendix:derivation_indep_posteriors}
\begin{align*}
    p(\rvf| \mathcal{D}) &= \frac{1}{p(\mathcal{D})}p(\rvf)\prod^M_{m=1} p(\vy_m|\rvf_{\mX_m})\\
    &= \frac{1}{p(\mathcal{D})} p_0(\rvf_\mZ)p(\rvf_*|\rvf_\mZ)\prod^M_{m=1}p(\rvf_{\mX_m}|\rvf_{\mZ_m})p(\rvy_m|\rvf_{\mX_m})\\
     &= \frac{1}{p(\mathcal{D})} p(\rvf_*|\rvf_\mZ)\prod^M_{m=1}p_0(\rvf_{\mZ_m})\prod^M_{m=1}p(\rvf_{\mX_m}|\rvf_{\mZ_m})p(\vy_m|\rvf_{\mX_m})\\
    &= \frac{1}{p(\mathcal{D})} p(\rvf_*|\rvf_\mZ)\prod^M_{m=1}p_0(\rvf_{\mZ_m})p(\rvf_{\mX_m}|\rvf_{\mZ_m})p(\vy_m|\rvf_{\mX_m})\\
    &= \frac{1}{p(\mathcal{D})} p(\rvf_*|\rvf_\mZ)\prod^M_{m=1}p_0(\rvf_{\mZ_m},\rvf_{\mX_m})p(\vy_m|\rvf_{\mX_m})\\
    &= \frac{p(\rvf_*|\rvf_\mZ)\prod^M_{m=1}p_0(\rvf_{\mZ_m},\rvf_{\mX_m}, \vy_m)}{\int p(\rvf_*|\rvf_\mZ)\prod^M_{m=1}p_0(\rvf_{\mZ_m},\rvf_{\mX_m}, \vy_m)d\rvf}\\
    &= \frac{p(\rvf_*|\rvf_\mZ)\prod^M_{m=1}p(\rvf_{\mZ_m},\rvf_{\mX_m}, \vy_m)}{\prod^M_{m=1}p(\mathcal{D}_m)}\\
    &= p(\rvf_*|\rvf_\mZ)\prod^M_{m=1}p(\rvf_{\mZ_m},\rvf_{\mX_m}| \vy_m)
\end{align*}

\section{BPC-fKL pseudocode (from \cite{Kim2022BPCDiv})}\label{sec:appendix:bpc-fklpseudocode}
\begin{algorithm}[H]
    \centering
    \caption{BPC-fKL}\label{alg:bpcfkl}
    \begin{algorithmic}[1]
        \STATE {\bfseries Require:} Set of expert trajectories $\mathcal{T}$, trained on $\mathcal{D}$, each parameter trajectory saves parameters to disk at a specified interval
        \STATE {\bfseries Require:} Number of SGD sampler steps with the BPC $L_\mathcal{C}$ and the data $L_\mathcal{D}$, total training steps $K$, number of Gaussian samples $S$, variances $\Sigma_\mathcal{C}$, $\Sigma_\mathcal{D}$, sampling learning rate $\eta$ and BPC learning rate $\psi$.
        \STATE {\bfseries Input:} local data $\mathcal{D} = (\mX, \vy)$
        \STATE Initialize $\mathcal{C} = (\mZ, \hat{\vy})$ from $\mathcal{D}$
        \FOR{$k \in \{1, ..., K\}$}
        \STATE Sample expert trajectory $\tau = \{\vtheta^{(r)}_*\}^T_{r=0}$ from trajectory set $\mathcal{T}$
        \STATE Randomly choose a checkpoint to start $r \leq L_\mathcal{D}$ and initialize $\vtheta^{(0)}_\mathcal{C} = \vtheta^{(0)}_\mathcal{D} = \vtheta^{(r)}_*$
        \STATE Obtain $\vtheta_\mathcal{D}^{(L_\mathcal{D})} = \vtheta_*^{(r + L_\mathcal{D})}$ from memory
        \FOR{$t \in \{1, ..., L_\mathcal{C}\}$}
        \STATE Update coreset parameter $\vtheta^{(t)}_\mathcal{C} \leftarrow \vtheta_\mathcal{C}^{(t-1)} - \eta \nabla_\vtheta \log p(\hat{\vy}|f(\mZ, \vtheta_\mathcal{C}^{(t-1)}))$
        \ENDFOR
        \STATE Sample random Gaussian noises $\{\epsilon^{(s)}_\mathcal{D}, \epsilon^{(s)}_\mathcal{C}\}^S_{s=1} \overset{i.i.d}{\sim} \mathcal{N}(0, \sigma_\epsilon)$
        \STATE $g \leftarrow \frac{1}{S} \sum^S_{s=1} \left( \nabla_\mathcal{C}\log p(\hat{\vy}|f(\mZ, \vtheta_\mathcal{D}^{(L_\mathcal{D})} + \Sigma^{1/2}_\mathcal{C}\epsilon^{(s)}_\mathcal{D})) - \nabla_\mathcal{C}\log p(\hat{\vy}|f(\mZ, \vtheta^{(L_\mathcal{C})}_\mathcal{C} + \Sigma^{1/2}_\mathcal{C}\epsilon^{(s)}_\mathcal{C})) \right)$
        \STATE $\mathcal{C} \leftarrow \mathcal{C} - \psi g$
        \ENDFOR        
        \STATE {\bfseries{Return} $\mathcal{C}$}
    \end{algorithmic}
\end{algorithm}

\section{Experiment details}\label{sec:appendix:experimentdetails}
\subsection{Datasets and models}\label{sec:appendix:datasets}
\textbf{Synthetic regression}
We generated data on a domain defined on 3 intervals: $\{(-0.8, -0.6), (-0.2, 0.0), (0.5, 0.8)\}$ using a ground truth function $y(x) = 1.5\sin(0.4 \pi x) + 1.5\cos(2\pi x) + \epsilon$ where $\epsilon \sim \mathcal{N}(0, 0.3)$. We generate 5 clients that sample datapoints from these intervals and compute outputs with $y(x)$. We introduce non-iid-ness to client datasets by sampling a probability vector over the 3 intervals from a Dirichlet distribution with $\alpha = [1.0, 1.0, 1.0]$. Each client thus samples more from specific intervals than other clients, which introduces the non-iid-ness in the data. We use a 3 layer MLP with 128 units per linear layer with Swish activation functions as neural network architecture.

\textbf{2d classification}
The dataset is the 'moons' dataset from the \textit{scikit-learn} library\footnote{https://scikit-learn.org/}. We add Gaussian noise with a standard deviation of $0.14$ to all samples. We generate 5 clients that then sample a dataset of only 20 points from the moons dataset. As a network architecture we use a 3 layer MLP with 50 units and relu activations, combined with Group Normalization with two groups in the first two layers \cite{wu2018group}.

\textbf{EMNIST-62}
The original EMNIST-62 dataset is an extended version of the classic MNIST dataset to 62 classes, adding both capital and undercase letters. The authors of LEAF partitioned this dataset into a federated dataset based on the author identification that came with the original data collection of EMNIST \cite{caldas2018leaf}. This gives the federated EMNIST-62 dataset a natural non-iid-ness between clients which allows for more realistic simulations, as the difference between author handwriting styles is a good example of the type of non-iid-ness that one could expect to see when implementing FL on real world data. As a network architecture we adopted the ConvNet architecture from \cite{reddi2021adaptivefo}, but without Dropout, as we found that negatively affected our method's performance.

\subsection{BPC-FL Hyperparameters}\label{sec:appendix:hyperparameters}
We provide all relevant hyperparameters for BPC-FL for each experiment in Table \ref{tbl:hyperparams}

\begin{table}
\centering
\begin{tabular}{|c|c|c|c|}
\hline
\textbf{Hyperparameter}                      & \textbf{Synthetic Regression} & \textbf{2d classification} & \textbf{EMNIST-62} \\ \hline
batch size                                   & 20                            & 50                         & 25                 \\
client update steps                          & 400                           & 700                        & 500                \\
prior precision / regularization                             & 1e-2                          & 1e-1                       & 1e-2               \\
random init std $\sigma_z$                              & 1.0                           & 6e-1                       & 1e-3               \\
step size~$x$                            & 0.01                          & 1e-3                       & 1e2                \\
step size~$y$                            & 1.0                           & 0.0                        & 0.0                \\
weight mc samples per update                                 & 10                            & 20                         & 5                  \\ \hline
\textbf{Client local hyperparameters}        &                               &                            &                    \\ \hline
number of pretrain seeds             & 100                           & 100                        & 100                \\
number of pretrain trajectories              & 100                           & 100                        & 100                \\
pretrain save interval                       & 5                            & 5                          & 5                  \\
number of pretrain steps                     & 300                           & 400                        & 750                \\
pretrain step size                       & 1e-2                          & 1e-2                       & 1e-2               \\
mcmc sampler                                 & Adam                          & Adam                       & SGD                \\
mcmc sampler step size                      & 1e-2 & 1e-2 & 1e-2\\
pseudodata optimizer                            & SGD                          & SGD                       & SGD                \\
data mcmc chain length                       & 150                           & 150                        & 200                \\
pseudodata mcmc chain length                    & 200                           & 200                        & 250                \\
number of noise samples $S$                    & 10                            & 10                         & 10                 \\
noise std  $\sigma_\epsilon$                                  & 1e-2                          & 1e-2                       & 1e-4               \\ \hline
\end{tabular}
\label{tbl:hyperparams}
\caption{Used hyperparameters for experiments}
\end{table}

\subsection{Baselines details}\label{sec:appendix:baselines}
\textbf{FedAvg} We use the adaptive implementation of FedAvg from \cite{karimireddy2021breaking} implemented in FedJAX. For EMNIST-62 we used the optimal hyperparameters from that paper. For the synthetic datasets we performed a grid search over $\{1e-3, 1e-2, 1e-1\}, \{1e-2, 1e-1, 1e0\}$ for the client and server step sizes, respectively. All hyperparameters are summarized in \ref{tbl:hyperparams_fedavg}.
\begin{table}[h]
\centering
\begin{tabular}{|c|c|c|c|}
\hline
\textbf{Hyperparameter}       & \textbf{Synthetic regression} & \textbf{2d classification} & \textbf{EMNIST-62} \\ \hline
Client optimizer     & Adam                 & Adam              & SGD       \\
Client step size & 1e-2                 & 1e-2              & 1e-1      \\
Server optimizer     & Adam                 & Adam              & SGD       \\
Server step size & 0.1                  & 0.1               & 1.0       \\ \hline
\end{tabular}
\label{tbl:hyperparams_fedavg}
\caption{Hyperparameters FedAvg baseline}
\end{table}

\textbf{DOSFL} We implemented the DOSFL algorithm ourselves following the details in \cite{zhou2020distilled}. We found that backpropagating through optimization trajectories made the algorithm too memory demanding for our resources, thus we could not produce results on the larger EMNIST-62 task. When setting the hyperparameters to different configurations that match the communication load of our method, DOSFL did not produce a useful server model. Therefore, we gradually increased distillation steps and epochs until a reasonable result was produced by the algorithm. We found that random masking and soft resetting did not improve performance. We also were not able to produce a good result for DOSFL on the synthetic regression task for any hyperparameter configuration that was within our memory constraints. The hyperparameters used for producing the results in the paper are given in Table \ref{tbl:hyperparams_dosfl}. 
\begin{table}[h]
\centering
\begin{tabular}{|c|c|c|}
\hline
\textbf{Hyperparameter}          & \textbf{Synthetic regression} & \textbf{2d classification}                                  \\ \hline
$x$ step size       & 0.1                  & 0.01                                               \\
$x$ optimizer           & SGD ($m=0.9$)        & SGD ($m=0.9$)                                      \\
$y$ step size       & 0.1                  & 0.0                                                \\
$y$ optimizer           & SGD ($m=0.9$)        & SGD ($m=0.9$)                                      \\
$\eta$ step size    & 0.0                  & 0.0001                                             \\
$\eta$ optimizer        & SGD                  & Adam ($\beta_1=0.9, \beta_2=0.999, \epsilon=1e-8$) \\
$\eta$ initial value     & 0.01                 & 0.01                                               \\
distillation steps      & 20                   & 30                                                 \\
distillation epochs     & 1                    & 20                                                 \\
distillation batch size & 10                   & 30                                                 \\ \hline
\end{tabular}
\label{tbl:hyperparams_dosfl}
\caption{Hyperparameters for DOSFL baseline}
\end{table}

\textbf{FedPA} We implemented the Federated Posterior Averaging (FedPA) algorithm ourselves, following mostly following the author's implementation \footnote{https://github.com/alshedivat/fedpa}. We used the paper's suggested hyperparameters for the EMNIST-62 learning task \cite{al-shedivat2021federated}. Interestingly, FedPA really tended to overfit on training clients when the training client pool was relatively small.

\textbf{MIME} We used the MIME implementation in FedJAX written by the paper authors. Interestingly, we did not find MIME to perform according to the reported results in \cite{karimireddy2021breaking} when using the corresponding hyperparameters. MIME typically performed worse on EMNIST-62 when we increased the amount of local steps.

\begin{table}[h]

      \centering
       \begin{tabular}{|l|l|}
        \hline
        \textbf{Hyperparameter} & \textbf{EMNIST-62} \\ \hline
        client optimizer        & SGD ($m=0.9$)      \\
        client step size    & 0.01               \\
        server optimizer        & SGD ($m=0.9$)      \\
        server step size    & 0.5                \\
        number of samples       & 10                 \\
        burn in rounds          & 500                \\
        shrinkage               & 0.1                \\ \hline
        \end{tabular}
        \caption{FedPA hyperparameters}
        \label{tbl:hyperparams_fedpa}
\end{table}
\begin{table}[h]
      \centering
        \begin{tabular}{|l|l|}
        \hline
        \textbf{Hyperparameter} & \textbf{EMNIST-62}                                 \\ \hline
        base optimizer          & Adam ($\beta_1=0.9, \beta_2=0.999, \epsilon=1e-8$) \\
        base step size      & 0.001                                              \\
        server optimizer        & SGD                                                \\
        server step size    & 1.0                                                \\
        gradient batch size     & 128                                                \\ \hline
        \end{tabular}
        \caption{MIME hyperparameters}
        \label{tbl:hyperparams_mime}
\end{table}
\newpage
\subsection{Downstream evaluation}\label{sec:appendix:eval}
We searched hyperparameters for the downstream optimizers and samplers by running Bayesian Optimization search over a fixed learned pseudodataset for a specific task. The final hyperparameters are shown below in table \ref{tbl:hyperparams_downstream}.
\begin{table}[h]
\centering
\begin{tabular}{|l|l|l|l|} 
\hline
\textbf{SGD hyperparameters}  & \textbf{Synthetic regression}        & \textbf{2d classification}            & \textbf{EMNIST-62}                   \\ 
\hline
step size                 & -                                    & 2e-2                                  & 1e-2                                 \\
number of steps               & -                                    & 500                                   & 2000                                 \\ 
\hline
\textbf{Adam hyperparameters} &                                      &                                       &                                      \\ 
\hline
step size                 & 1e-2                                 & -                                     & 1e-2                                 \\
$\beta_1$   & 0.9                                  & -                                     & 0.9                 \\              $\beta_2$   & 0.999                                & -                                     & 0.999                                \\
$\epsilon$   & 1e-8                                 & -                                     & 1e-8                                 \\
number of steps               & 300                                  & -                                     & 2000                                 \\ 
\hline
\textbf{HMC hyperparameters}  &                                      &                                       &                                      \\ 
\hline
step size                     & 1e-3                                 & 2e-3                                  & 2e-4                                 \\
inverse mass matrix           & $\text{diag}(50)$ & $\text{diag}(100)$ & $\text{diag}(10)$  \\
num integration steps         & 20                                   & 30                                    & 10                                    \\
number of steps               & 200                                  & 40                                    & 150                                   \\
number of samples             & 40                                   & 50                                    & 10                                   \\
\hline
\end{tabular}
\label{tbl:hyperparams_downstream}
\caption{Hyperparameters for downstream inference algorithms on evaluation coreset}
\end{table}

\subsection{Additional results}\label{sec:appendix:additional}

\subsubsection{Synthetic regression, small datasets}
\begin{figure}[h]
    \centering
    \includegraphics[scale=0.18]{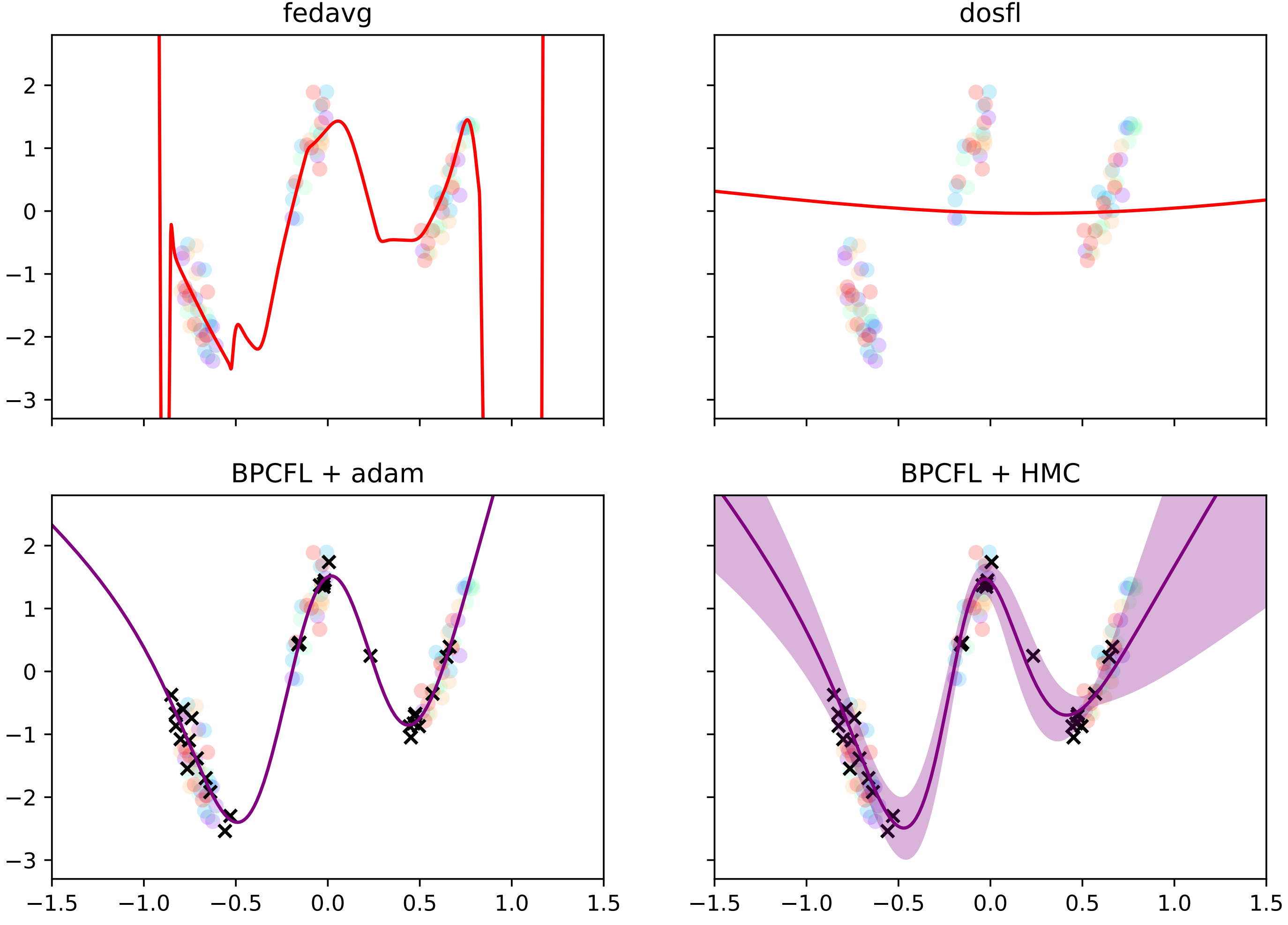}
    \caption{Synthetic regression - small datasets}
    \label{fig:regression-small data}
\end{figure}

Figure \ref{fig:regression-small data} shows the synthetic regression problem for a setting where clients have very little data. We simulated 5 clients that only have 20 datapoints. BPC-FL inferred a pseudodataset of 6 datapoints as before. It is interesting to see that FedAvg starts to produce functions that overfit when clients have smaller datasets, while BPC-FL still produces decent results.

\subsubsection{EMNIST62 - influence of amount of pseudodata size}
In Table \ref{tbl:coresetsize} we show the final performance of models produced by SGD after running BPC-FL on 100 clients for different BPC sizes. Note how performance improves with a larger amount of pseudodatapoints. We did observe that for larger pseudodatasets, pseudoinputs started to exactly mimic $\vx$-values of datapoints with only a few classes in the dataset.
\begin{table}[h]
\centering
\begin{tabular}{|c|c|c|c|}
\hline
\textbf{Number of pseudo-input}       & \textbf{Negative log likelihood} & \textbf{Accuracy} & \textbf{ECE} \\ \hline
30     & $1.504 \pm 0.150$                 & $0.703 \pm 0.005$             & $0.111 \pm 0.040$       \\
62 &     $1.118 \pm 0.051$             &  $0.720 \pm 0.003$              & $0.080 \pm 0.027$      \\
100     & $1.00 \pm 0.03$                 & $0.755 \pm 0.004$              & $0.064 \pm 0.023$       \\ \hline
\end{tabular}
\label{tbl:coresetsize}
\caption{Influence of pseudodata size on performance}
\end{table}

\subsubsection{EMNIST62 - influence of number of clients}
In Table \ref{tbl:num clients}, we show the final performance of models produced by SGD after running BPC-FL on different amount of clients, with a pseudodataset size of 62. It is shown that for more clients that participate, the quality of the server model improves, though it seems that after 100 clients the performance increase starts to stall. We suspect that because we learn a compressed representation of client information, some fine-grain information in the data posterior will go lost, which will not be gained back by simply adding more clients that also carry that approximation error.
\begin{table}[h]
\centering
\begin{tabular}{|c|c|c|c|}
\hline
\textbf{Number of clients}       & \textbf{Negative log likelihood} & \textbf{Accuracy} & \textbf{ECE} \\ \hline
30                    & $1.822 \pm 0.132$             & $0.613 \pm 0.009$   & $0.150 \pm 0.004$   \\
50          &  $1.385 \pm 0.006$              & $0.651 \pm 0.006$     & $ 0.111 \pm 0.009$\\
100              &  $1.210 \pm 0.029$              & $0.712 \pm 0.103$    & $0.083 \pm 0.005$ \\
200              &  $1.152 \pm 0.073$              & $0.719 \pm 0.003$    & $0.092 \pm 0.031$ \\
300                & $1.01 \pm 0.002$              & $0.718 \pm 0.005$    &  $0.071 \pm 0.009$ \\ \hline
\end{tabular}
\label{tbl:num clients}
\caption{Influence of the amount of clients on performance}
\end{table}

\textbf{EMNIST62 $\mZ$ examples}
To give an idea of what the learned pseudodata $\mZ$ look like visually for the EMNIST-62 dataset, we provide some examples below.
\begin{figure}[h]
    \centering
    \includegraphics[scale=0.5]{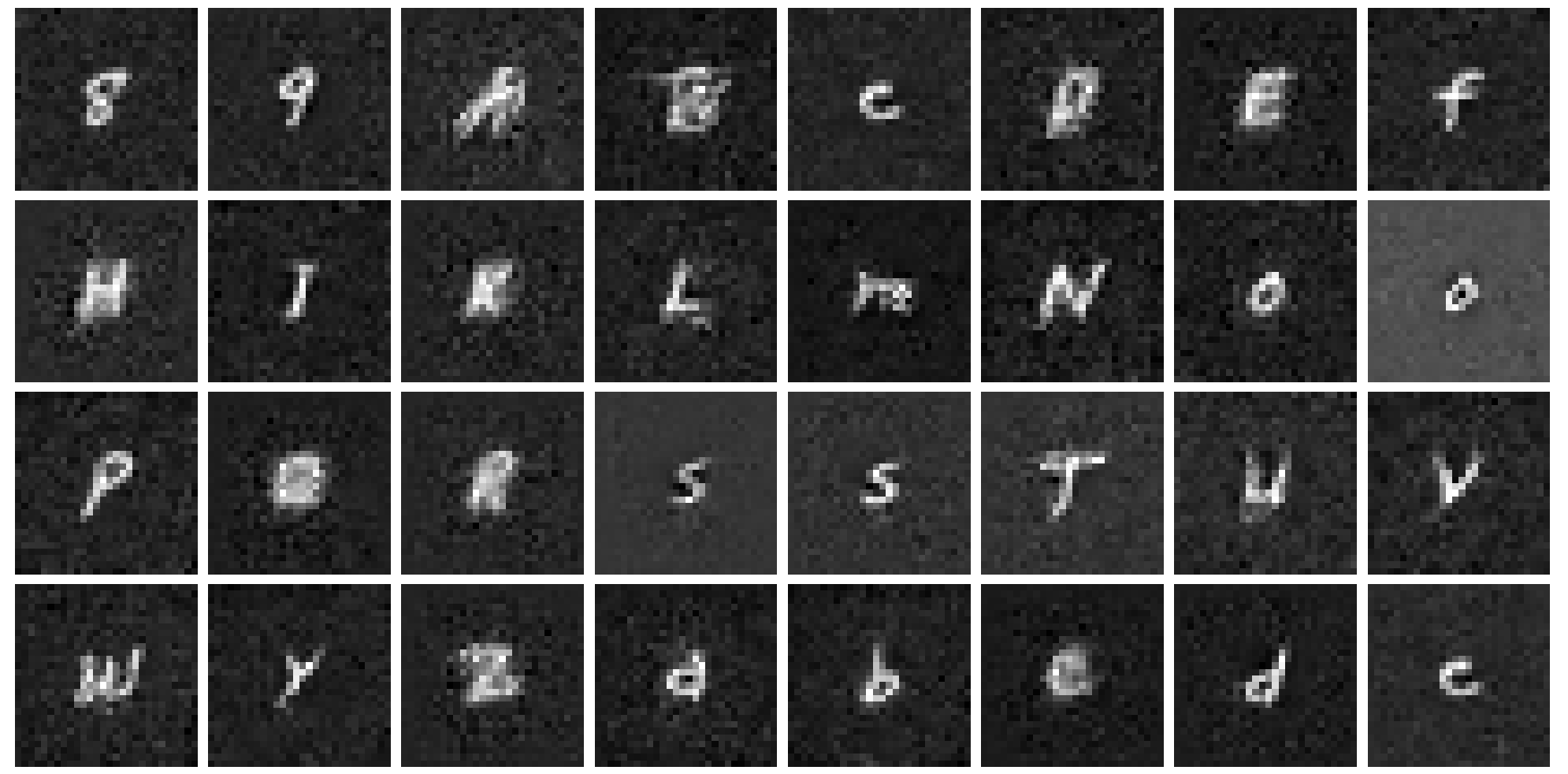}
    \caption{Some example x-values from the server set in EMNIST-62}
    \label{fig:example images}
\end{figure}